# Triaging moderate COVID-19 and other viral pneumonias from routine blood tests

Forrest Sheng Bao, PhD[1,*], Youbiao He[1,*], Jie Liu, MD, PhD[2,*], Yuanfang Chen[3,4], Qian Li, MD[5], Christina R. Zhang, MD[6], Lei Han, MD, PhD[4,7], Baoli Zhu[4,7], Yaorong Ge, PhD[8], Shi Chen, PhD[9,11#], Ming Xu, MD, PhD[4, 7, 9,#], Liu Ouyang, MD, PhD[10,#]

1. Department of Computer Science, Iowa State University, Ames, IA, 50011, USA
2. Department of Radiology, Union Hospital, Tongji Medical College, Huazhong University of Science and Technology, Wuhan 430022, China
3. Institute of HIV/AIDS/STI Prevention and Control, Jiangsu Provincial Center for Disease Control and Prevention, Nanjing 210009, China
4. Public Health Research Institute of Jiangsu Province, Nanjing 210009, China
5. Department of Pediatrics, Kunshan People's Hospital Affiliated To Jiangsu University, Kunshan 215300, China
6. Division of Neurology, Department of Medicine, University of British Columbia, Vancouver, BC V6T 1Z3, Canada
7. Department of Occupational Disease Prevention, Jiangsu Provincial Center for Disease Control and Prevention, Nanjing 210009, China
8. Department of Software and Information Systems, University of North Carolina at Charlotte, Charlotte, NC 28223, USA
9. Department of Public Health Sciences, College of Health and Human Services, University of North Carolina at Charlotte, Charlotte, NC 28262, USA
10. Department of Orthopaedics, Union Hospital, Tongji Medical College, Huazhong University of Science and Technology, Wuhan 430022, China
11. School of Data Science, College of Health and Human Services, University of North Carolina at  Charlotte, Charlotte, NC 28262, USA

Emails: fsb@iastate.edu, yh54@iastate.edu, liu_jie0823@163.com, yf_chen2010@163.com, qianyingbingfeng@163.com, christina.zhang@alumni.ubc.ca, hanlei@jscdc.cn, zhubl@jscdc.cn, yge@uncc.edu, schen56@uncc.edu, sosolou@126.com, ouyangliu211@hust.edu.cn

**Abstract**: The COVID-19 is sweeping the world with deadly consequences. Its contagious nature and clinical similarity to other pneumonias make separating subjects contracted with COVID-19 and non-COVID-19 viral pneumonia a priority and a challenge. However, COVID-19 testing has been greatly limited by the availability and cost of existing methods, even in developed countries like the US. Intrigued by the wide availability of routine blood tests, we propose to leverage them for COVID-19 testing using the power of machine learning. Two proven-robust machine learning model families, random forests (RFs) and support vector machines (SVMs), are employed to tackle the challenge. Trained on blood data from 208 moderate COVID-19 subjects and 86 subjects with non-COVID-19 moderate viral pneumonia, the best result is obtained in an SVM-based classifier with an accuracy of 84%, a sensitivity of 88%, a specificity of 80%, and a precision of 92%. The results are found explainable from both machine learning and medical perspectives. A privacy-protected web portal is set up to help medical personnel in their practice and the trained models are released for developers to further build other applications. We hope our results can help the world fight this pandemic and welcome clinical verification of our approach on larger populations.

# 1. Introduction

COVID-19 is a pandemic that has devastated the lives of billions of people around the world in recent months [1], causing hundreds of thousands of deaths worldwide and overwhelming hospitals in hot spots like Lombardy, Italy or New York City, USA. COVID-19, at its early stage, shares many symptoms with other illnesses, especially other virus-induced pneumonias, such as coughing and fevering [2][3]. Because of its extremely contagious nature, it is important to quickly distinguish COVID-19 from other viral pneumonias, in order to facilitate safe isolation of COVID-19 patients.

The gold standard to diagnose COVID-19 is genome testing. But it is limited due to accessibility and cost. No other ways can better diagnose or rule out COVID-19. For example, COVID-19 has very nonspecific lung CT imaging findings [4], which are similar to findings seen in other viral pneumonias, including those secondary to influenza [5]. Moreover, especially in developing countries ravaged by the virulent COVID-19, CT imaging, genomic, and even antibody/serology (e.g., IgM/IgG) testing are all likely to be poorly accessible. Even in the United States, there is still a huge gap between testing capacity and testing needs [6]. By late April, many Americans are still denied from testing due to issues like the shortage of swabs [7].

Therefore, in this paper, we turn our attention to a commonly accessible modality: routine blood tests [8], hoping to distinguish COVID-19 from other viral pneumonias based on widely-available, common blood chemistry profiles. Compared with other testing methods, routine blood tests have much more available supplies, equipment, and personnel. Massively and pervasively performed at hospitals and labs daily, they can be very affordable. But as we will show in this paper, this task is not trivial due to complicated underline correlation between blood biomarkers. Hence, a powerful tool to mine complex patterns from data is needed.

As a subarea of Artificial Intelligence (AI), machine learning (ML) studies giving computers a skill (usually from data) automatically without being manually programmed to do so. Hence, we hope to leverage ML methods to make use of routine blood tests for differentiating moderate COVID-19 cases from other moderate viral pneumonia cases automatically. Thus, given blood test values of a moderate viral pneumonia patient, the computer is expected to tell whether s/he is contracted with COVID-19 or other viruses. An ML method acquires such a skill through learning a mathematical model from blood test values of many human subjects and whether they are contracted with COVID-19. The reason we focus on moderate cases is because a great majority of COVID-19 patients show moderate symptoms, and they are contagious although most of them eventually recover on their own. Once detected early, such patients can be isolated to cut the transmission chain.

The ML problem in this paper is a typical classification task. To solve, we employ random forests (RFs) [9] and support vector machines (SVMs) [10], two families of classifiers that have been widely and empirically considered robust and powerful [11]–[13]. Our experimental results show that the SVM-based approach is superior to the RF-based in general. An RBF-kernel SVM trained on 294 subjects, 208 COVID-19 vs. 86 non-COVID-19, achieves an accuracy of 84%, a sensitivity of 88%, a specificity of 80%, and a precision of 92% in differentiating COVID-19 from non-COVID-19 viral pneumonia at moderate stage. Thorough medical explanations can justify the effectiveness of using ML to tackle the problem.

We completely open our results to help the medical practice and research to fight this pandemic. Our model is released at http://github.com/forrestbao/covid-19. We further release a privacy-protected web portal based on the trained model to assist medical workers to triage patients at http://forrestbao.github.io/covid-19 All source code will be released upon the acceptance of this manuscript at a peer-reviewed journal. Although the results are promising, we look forward to verifying its effectiveness on larger populations with researchers and doctors around the world.

## 2. Data and problem formulation

### 2.1 Subjects
We treat a blood **test** in the medical context, such as white blood cell count or the concentration of a specific protein in serum, as a **feature** in the machine learning context. Some blood tests are often ordered together, forming a blood test panel or blood test work. This study is based on routine blood tests from 3 groups (called **classes** in machine learning) of subjects:
1. Patients with **moderate non-COVID-19 viral** pneumonia (N = 86, age: 50.65 ± 17.38, mean 52; gender: 30 males and 56 females), short as "**viral**"
2. Patients with **moderate** COVID-19 (N = 208, age: 47.96 ± 15.05, mean 47, gender: 85 males and 123 females), short as "**moderate**"
3. Patients with **severe** COVID-19 (N = 118, age: 67.57 ± 11.94, mean 68, gender: 62 males and 56 females), short as "**severe**"

The lab tests were ordered independently by individual doctors in two hospitals in China: Wuhan Union Hospital, in Wuhan where the outbreak started, and Kunshan People's Hospital, in Kunshan, which is 1 hour west of Shanghai. All COVID-19 data from Wuhan were collected between Jan. 18 and Feb. 22, 2020 when patients were initially admitted into the hospital, while all non-COVID-19 blood tests from Kunshan were performed between Jan. 1, 2016 and Dec. 1, 2019. The temporal separation minimizes the chance of mixing data of COVID-19 subjects and non-COVID-19 subjects. All COVID-19 subjects were confirmed with RT-PCR testing of SARS-COV-2 (the virus causing COVID-19) genome. All non-COVID-19 subjects were confirmed with radiology findings and had no bacterial infections. All COVID-19 subjects had no bacterial infections.

Note that the distinction between moderate and severe conditions is a clinical diagnosis and hence subjective. A moderately trained doctor can differentiate between the two easily. Usually a patient with severe pneumonia will be overall sicker, e.g., with a lower SpO2, or unstable/ abnormal vital signs. According to the COVID-19 guideline by National Health Commission of China [14], a severe adult subject is who meets any one of the following:
1. Shortness of breath, RR > 30 breaths/minute;
2. Oxygen saturation < 93% at rest
3. Arterial oxygen partial pressure (PaO2)/ fraction of inspired oxygen (FiO2) < 300mmHg (1mmHg=0.133kPa).

Any subject less severe than severe but with radiological findings of pneumonia is considered moderate. Note that in the Chinese guideline, there is one more level above severe called critical, for anyone needing mechanical ventilation or ICU, or having shock. They are excluded from this study. All subjects in this study are adults.

## 2.2 Features

Because the blood tests were ordered by different doctors over a long span of time, not all subjects had identical features (again, a feature is a blood test), and nearly no feature has values on all subjects. To deal with the sparsity of the raw data, we first excluded any feature missing on half of the initial subject pool in any group, and then removed any subject missing more than 20% of features. The initial pools of 99 viral, 213 moderate, and 122 severe subjects were thus narrowed down to 86 viral, 208 moderate, and 118 severe subjects, respectively. The demographics of final subjects were given in **Section 2.1** above. The features were narrowed down to those in **Table 1**. There were some additional common features between severe COVID-19 and moderate COVID-19 detailed in **Table 2**.

**Table 1**: Features shared among the 3 classes

| Feature/test name (abbreviations) | unit |
|---|---|
| Count of White blood cell (WBC) | Million cells per liter ($10^9$/L) or thousand cells per microliter ($10^3$/mL) |

| Hemoglobin (HGB) | grams per liter (g/L) |
|---|---|
| Platelet count | Million cells per liter ($10^9$/L) or thousand cells per microliter ($10^3$/mL) |
| Neutrophil percent | % |
| Neutrophil count | Million cells per liter ($10^9$/L) or thousand cells per microliter ($10^3$/mL) |
| Lymphocyte percent | % |
| Lymphocyte count | Million cells per liter ($10^9$/L) or thousand cells per microliter ($10^3$/mL) |
| C-reaction protein (CRP) | Milligrams per liter (mg/L; to convert to mg/dL, divide by 10) |
| Total bilirubin (TBL) | Micromole per liter (umol/L; to convert to mg/dL, divide by 17.1036) |
| Blood urea nitrogen (BUN) | millimole per liter (mmol/L; to convert to mg/dL, divide by 0.3571) |
| Creatinine | Micromole per liter (umol/L; to convert to mg/dL, divide by 88.417) |
| Lactate dehydrogenase (LDH) | Units per liter (U/L) |
| D-dimer | micrograms per liter (mg/L; to convert to ng/mL, divide by 0.001) |

**Table 2**: Additional features for severe vs. moderate

| Feature name (abbreviations) | Unit |
|---|---|
| IL-6 | pg/ml |
| Erythrocyte sedimentation rate (ESR) | Millimeters per hour (mm/h) |
| Procalcitonin | Nanograms per microliter (ng/mL) |
| Alanine transaminase (ALT) | Units per liter (U/L) |
| Aspartate transaminase (AST) | Units per liter (U/L) |
| Creatine kinase (CRK) | Units per liter (U/L) |

Several preprocessing steps were performed on the remaining data. A missing value on a feature was replaced with the mean of that feature. On each feature, all values were standardized to zero mean and unit variance. For c-reaction protein (CRP), when a high-sensitivity CRP (hsCRP) value was available, we used the hsCRP value as the CRP value.

Curious about how demographics play a role, we will report the results with and without consideration of gender and age separately. In the former case, age and gender become two additional features.

### 2.3 Tasks

The 3 groups of subjects thus form 3 binary classification tasks:
1. [primary] moderate vs viral (N=208 vs. 86)
2. [secondary] Severe vs. viral (N=118 vs. 86)
3. [bonus] Severe vs. moderate (N=117 vs. 204[1])

Medical workers need help the most from the primary task of differentiating moderate COVID-19 cases from cases of non-COVID-19 moderate pneumonia. The secondary task is not a very fair apple-to-apple comparison as the difference in blood test values could be a result of different pneumonia severities rather than different pathogens. The bonus task is introduced simply out of curiosity, because severe subjects can be easily distinguished from moderate subjects based on blood oxygen saturation and/or other symptoms such as difficulty to breath.

---

[1] The numbers are not 118 vs 208 because 5 samples have too many missing values in additional features in Table 2.

Mathematically, the goal of each of the classification problems above is finding a function $f : \mathbb{R}^d \to \mathbb{Z}$ representing a complex relationship between $d$ blood test values of a subject and the diagnosis of the subject. For example, for the primary task,
- $f(x_1, x_2, \cdots, x_d) = 1$ if the subject is contracted with moderate COVID-19, and
- $f(x_1, x_2, \cdots, x_d) = -1$ if with moderate non-COVID-19 viral pneumonia.

Of course, for all human subjects, the order of features must be identical. Per **Table 1**, in this paper, $x_1$ is always how many white blood cells (WBCs) per microliter, $x_2$ is always how many milligrams of CRP per liter, $x_4$ is always the percentage of neutrophil among all white blood cells, etc. In machine learning, the vector composed of all feature values $X = [x_1, x_2, \cdots, x_d]$ is called a **feature vector**, the function $f$ is called a **classifier**, and the output of $f$ is called a **label** which is one of the known classes.

The function $f$ is usually obtained empirically from **samples**, which are pairs of feature vectors and expected labels, through a process called **training**. Each sample corresponds to a human subject in this paper. The process of plugging in an (unseen) feature vector $X'$ into the classifier to yield an output $f(X')$ is called the **predicting**. A bad or poorly trained classifier would simply memorize all samples. When given an unseen sample, the classifier very likely will not output the expected label. A good classifier instead captures the pattern of the data across features holistically to maximize its ability to deal with radical (or extreme, outlying) unseen samples.

As we will see later in this paper, a classifier often involves many parameters that will not be changed by the training samples in the training process. Such parameters are called **hyperparameters** and need to be customized on a problem-by-problem basis to maximize the performance of the classifier.

## 3. Methods

Machine learning is needed for the tasks above because blood test features have rather complicated relationships or correlations which will be missed and not made use of by simple thresholding on raw features, e.g., claiming that CRP>2.5 means COVID-19, otherwise non-COVID-19 viral pneumonia. For example, in two studies of hospitalized COVID-19 patients, one [2] of which has more severe subjects than the other [15], while LDH elevation above its normal range is observed on similar ratios of patients (73% and 76%, respectively), WBC and lymphocyte decreases below their normal ranges are found on very different ratios of patients: 25% [2] vs. 9% [15] for WBC and 63% [2] vs. 35% [15] for lymphocytes. This shows that LDH elevation is more severity-invariant than WBC or lymphocyte count, and WBC and lymphocyte changes seem to be correlated. While thresholding on LDH might capture the characteristics of COVID-19, fixed or limited discrete thresholds on WBC and lymphocytes will be less effective. A rule that captures the joint change in WBC and lymphocytes might be more desired. Experimental results later also suggest that thresholding-based approaches, including decision trees and random forests, are outperformed by SVM-based which can build complicated functions involving multiple features through kernel methods.

We pick two empirically effective and robust families of classifiers, random forests (RFs) and support vector machines (SVMs), as representatives to study the general feasibility and effectiveness of using ML to make use of routine blood tests for COVID-19 triage. As the state of the art before deep learning [11], [12], another popular branch of ML, RFs and SVMs are considered suitable for problems with small amounts of samples. Our problem is not suitable for deep learning at this point due to the limited number of samples. Once much more data becomes available, it will definitely be worth trying deep learning-based approaches. In this section we will briefly go over RFs and SVMs, and graphically explain how they work to our audience who are not familiar with ML.

## 3.1 Decision trees and Random forests

Instead of developing an analytical form (e.g., an equation) of the function $f$, decision trees (DTs) [16] and random forests (RFs) [9] achieve classification by using binary comparison rules. A random forest is a collection of DTs. A DT, standalone or as part of an RF, is a set of hierarchical binary thresholding rules like a nested sequence of if-then-else statements. **Figure 1** is an example DT in a graphical representation. Each rectangular box is called a **node** on the tree and the tree starts with the **root** node at the top. Each node compares the value of a feature with a threshold, e.g., CRP <= 2.23 in the root node. Arrows represent the order of applying rules, e.g., if CRP>2.23 then we check whether creatinine is <= 48.65. A node that has no arrow pointing out of is called a **leaf** node, which represents a final classification decision. Given an unseen sample, starting from the root node, we compare the feature value with the threshold, and then go to the next node based on the result of comparison, until reaching a leaf node. The sequence of nodes visited from the root to the leaf node is called a **path**. The number of nodes on a path minus 1 is called the **depth** of the path. So the leftmost path in **Figure 1** means that if a subject's CRP <=2.23, and D-dimer<=1.1, and platelet count <= 359.0, then this subject is classified contracted with non-COVID-19 viral pneumonia. A feature may be checked more than once with different thresholds, e.g., D-dimer is checked twice on the leftmost path with moderate as the leaf node (CRP <=2.23, D-dimer<=1.1, platelet count> 35.9.0, and D-dimer<=0.97). The feature to be compared at each node, the order of comparison rules, and the thresholds are all obtained through training. The maximum depth of a path is a parameter to be searched using grid search in **Section 4.1**.

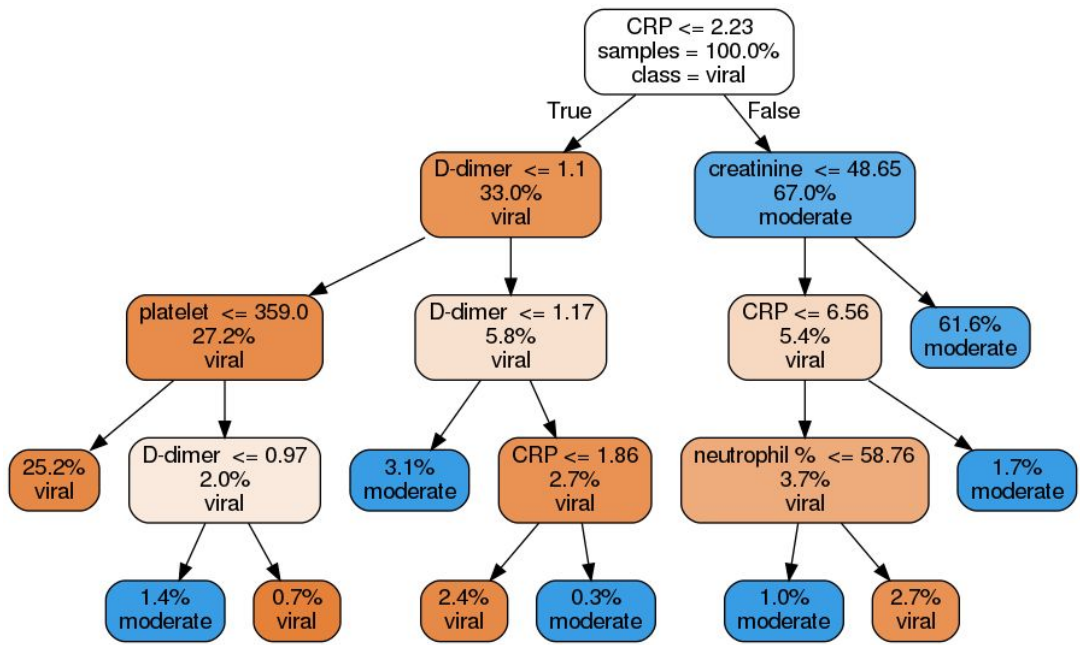

**Figure 1**: An example decision tree for classifying moderate COVID-19 vs. moderate non-COVID-19 viral from blood tests. For each non-leaf node block, the threshold is at the top, with the portion of samples falling into this node in the middle, and the dominant class at the bottom. For leaf nodes, the thresholds are gone. The most important feature is CRP, at the root (top) of the decision tree. By thresholding CRP at 2.23, samples below the threshold are dominantly moderate non-COVID-19 pneumonia while those above or equal to are dominantly moderate COVID-19.

A decision tree is constructed recursively from an empty tree. Starting from the root node, a feature and a threshold are chosen to branch or fork (technically **split**) a node into two. The recursive process to construct the tree stops when all paths end up at leaf nodes. Among many ways to determine the feature and the threshold, this paper employs one of the classics, the CART algorithm [16] whose core is the Gini impurity or Gini index.

For a Boolean comparison condition S, its Gini impurity is defined as: $g(S) = 1 - \sum_{j=\pm 1} P^2(class = j | S)$, where P is the probability density function. Gini impurity measures how likely a randomly chosen sample satisfying a condition would be incorrectly labeled if it was randomly labeled according to the label distribution among all samples satisfying that condition [17], [18]. Given a feature F and a threshold T, the expectation of Gini impurity over two mutually exclusive conditions, F>T and F<=T, is computed as $E_g(F, T) = P(F > T) g(F > T) + P(F \leq T) g(F \leq T)$. The CART algorithm exhaustively searches over all features and all their values to find the combination of F and T that maximizes $E_g(F, T)$, i.e., $\arg\max_{F,T} E_g(F, T)$, and split a node accordingly.

The sample pool over which the probabilities above are calculated changes in the iterative process. It begins with all samples when splitting the root node. Then for non-root nodes, the sample pool includes only those already satisfying the conditions along the path. For example, in **Figure 1**, to split the left node under the root (D-dimer<=1.1), the sample pool includes only those already satisfying the condition CRP<=2.23. Hence the probability calculation should reflect the population reduction as we go down the tree. When to stop the split and to create a leaf node depends on 3 hyperparameters: the maximum depth of a path, the minimum Gini impurity, and the minimum sample pool.

A random forest is just a collection (technically an **ensemble**) of DTs that share the same set of hyperparameters. Those member DTs are trained from different sets of samples randomly taken with replacement from the same set of original samples. And when constructing each node, not all features are considered but only a random subset. The member DTs make predictions independently and the final classification of the RF is the majority vote of all DTs. The number of member DTs and the size of the random subset of features are two hyperparameters to be searched using grid search in **Section 4.1**.

### 3.2 To threshold or not to threshold, this is a question

Now allow us to use a toy problem to graphically illustrate how a DT and an RF work, and explain when SVMs might be better. **Figure 2** visualizes samples, each of which has two features (i.e., d=2), using a 2D Cartesian coordinate system such that each axis represents a feature. The blue squares and red circles represent two classes (+1 and -1) of samples. Obviously the two classes have quite separate means along the X (horizontal) axis and only the tails of their distributions (along X axis) slightly overlap. Thresholding on the X axis, as depicted by the vertical black line, would perfectly separate the two classes, without using the information from Feature 2. This can be easily implemented using a DT of only one node that thresholds on Feature 1 and also splits the plane into the left and right halves.

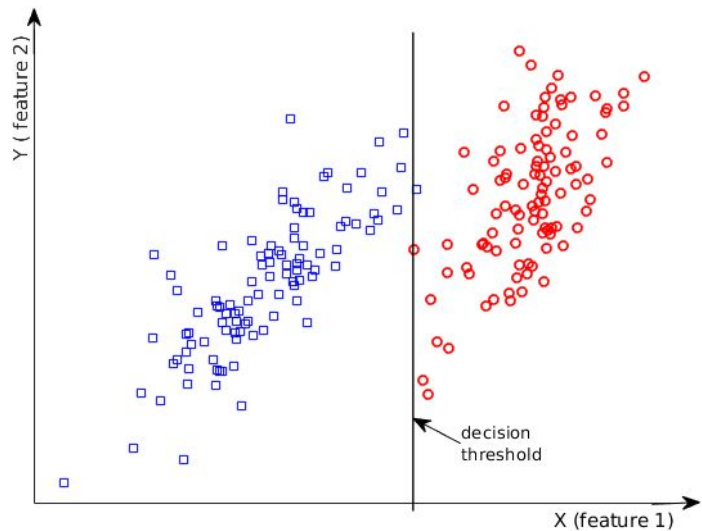

**Figure 2**: A toy classification problem. The distribution of samples here do not reflect that of our data.

An RF trained on the same dataset using both features will further split the plane into many tiles as shown in **Figure 3**. Recall that an RF uses majority votes of its composing DTs to classify. The darkness of a tile in **Figure 3** represents the extent of majority, e.g., the RF is most sure that samples distributed to the top right belong to class -1 (red) and those to the bottom left belong to +1 (blue).

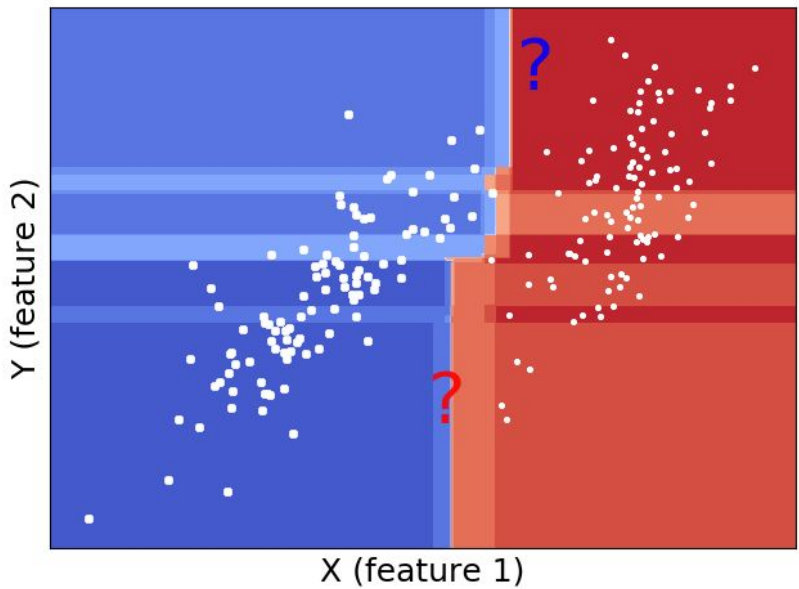

**Figure 3**: An example showing the decision boundary of a random forest-based classifier

But if we look at the overall picture, the two classes both exhibit a "diagonal" distribution. Unseen samples of class 1 (the big blue question mark) might appear to the top right side of the decision line while unseen samples of class -1 (the big red question mark) to the bottom left. So a sloped decision line, as illustrated by the black dash line in **Figure 4**, is more preferred for being more future-proof. The goal of a linear classifier, such as an SVM, is to find such a sloped decision line (called a **decision hyperplane** when multi-dimensional) that would not have been easily obtained by thresholding on single features. In our example, an RF can only achieve such a sloped decision hyperplane limitedly, e.g., the "staircase" pattern of red tiles in **Figure 3**.

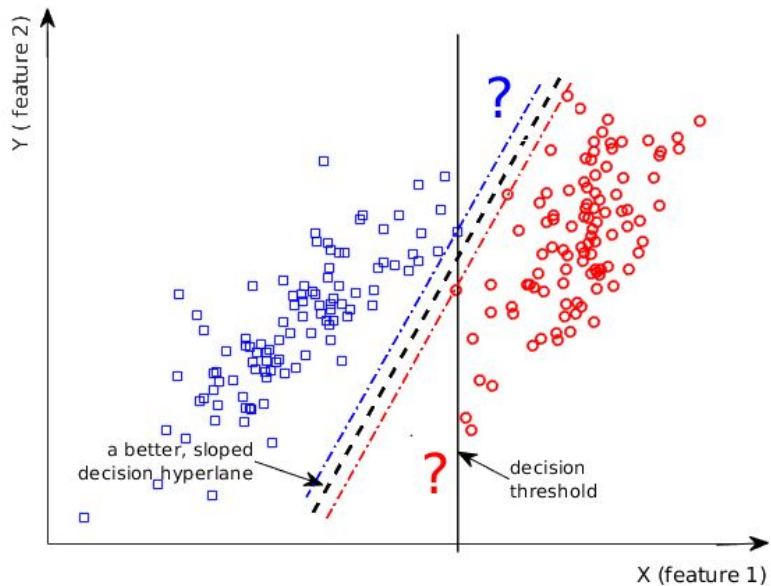

**Figure 4**: A sloped decision line more robust for potentially unseen samples depicted by big question marks.

To decide which sloped decision hyperplane is the best, SVMs introduce a concept called the **margin**, which is a strip- or slice-shaped zone (bounded by the blue and red dot-dash lines in Figure 1) centered at and parallel to the decision hyperplane and without any samples within. An SVM maximizes the width of the margin, or in other words, finds the widest "safe zone" between outliers of the two classes. A soft-margin SVM moves one step further by allowing a few samples (that make the envelope bumpy or zigzaggy) fall into the margin in exchange for a better slope to widen the margin. **Figure 5** illustrates a flatter decision hyperplane and its wider margin which a handful of samples fall into. SVMs use a constant C to control the tradeoff between margin width and samples fall into the margin (including misclassifications). A larger C means less tolerance to samples in the margin, while a smaller C means a wider margin. As we will see, SVMs can be easily extended further to achieve nonlinear decision hyperplanes.

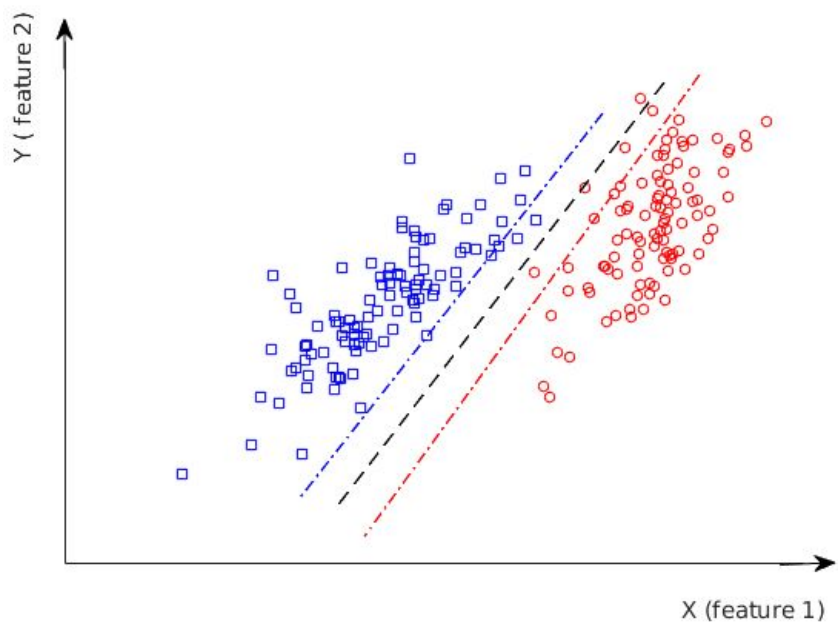

**Figure 5**: A soft-margin SVM with a wider margin tolerating a few samples falling into the margin.

## 3.3 Support Vector Machines (SVMs)

After the intuitive explanations above about SVMs, let us formally introduce it. Given samples in two classes, an SVM [10] (strictly speaking, a hard-margin linear SVM) finds a multi-dimensional direction along which the closest samples of both classes are apart the most. In other words, the difference among most alike samples of two classes are maximized. This idea gives SVMs great generalizability, making it superior to RFs on popular benchmarks [12]. SVMs have been theoretically and empirically proven to be adept and robust with a small amount of samples.

Without losing generality, an SVM-based classifier is defined as $f(X) = sign\,(W \cdot X + b)$ where $sign$ is the sign function, $X$ is the weight vector, the operation $\cdot$ is dot product between two vectors, and $b$ a scalar is the bias. Given a sample $X'$, the prediction is $f(X') = sign\,(W \cdot X' + b)$. The loss function is defined as $J(W) = \frac{\|W\|^2}{2} + C\sum_i \xi_i^2$ where the operator $\|\cdot\|$ means L2-norm, the hinge loss $\xi_i = \max\,(1 - y_i(W \cdot X_i + b), 0)$ represents the prediction error, $y_i$ is the ground truth label for the $i$-th sample $X_i$, and C is the hyperparameter mentioned above that controls the tradeoff between margin width and samples falling into the margin. An SVM can be derived into a dual formulation where the goal is to maximize the new loss function $J(\alpha_1, \alpha_2, \cdots) = \sum_i \alpha_i - \frac{1}{2}\sum_{i,j} \alpha_i \alpha_j y_i y_j X_i \cdot X_j$, subject to $\forall i,\ \alpha_i \geq 0$ and $\sum_i \alpha_i y_i = 0$. In the dual formulation, given a new sample $X'$, the prediction is $f(X') = sign\,(\sum_i \alpha_i y_i X_i \cdot X' + b)$. The SVMs discussed so far are linear SVMs, where the classification relies on linear relationships (a simple weighted sum) of features.

In order to exploit the nonlinear inter-feature relationship (e.g., BMI is a feature built on top of two features, weight and height) for a better performance, samples can be first (usually nonlinearly) mapped into a feature space via a mapping $\Phi$ before being fed into an SVM [19]. This results in a nonlinear SVM-based classifier $f(X') = sign\,(\sum_i \alpha_i y_i \Phi(X_i) \cdot \Phi(X') + b)$ where the dot product part is usually rewritten as a kernel function $K(X_i, X') = \Phi(X_i) \cdot \Phi(X')$. **Figure 6** illustrates such a transformation. One of the commonly used nonlinear kernels is the radial basis function (RBF) kernel $K(X_i, X') = \exp(-\gamma \|X_i - X'\|^2)$ where $\gamma$ is another hyperparameter. A nonlinear SVM can also be soft-margin, with samples falling into the margin. In **Section 4**, we will report results obtained in linear SVMs and RBF-kernel SVMs separately.

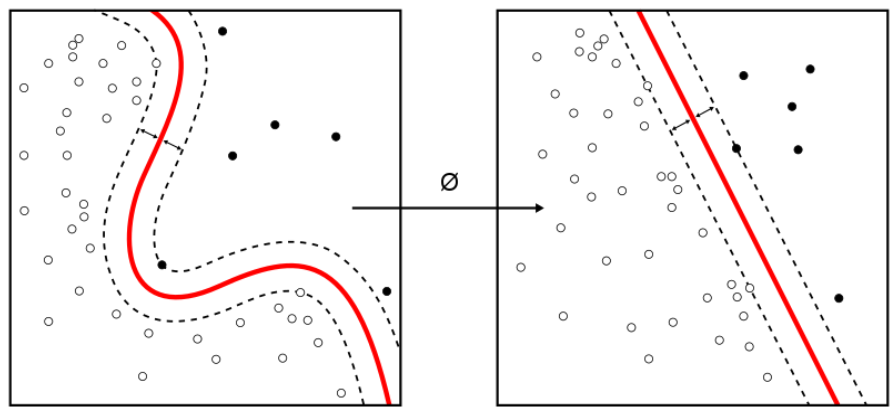

**Figure 6**: An example showing how to achieve nonlinearity in (hard-margin) SVMs.
Source: https://commons.wikimedia.org/wiki/File:Kernel_Machine.svg

# 4. Results and Discussions

## 4.1 Setups

**Cross validation**: Because the purpose of training a classifier is to use it to make predictions on unseen samples (e.g., given blood test values of a new subject, tell whether s/he is contracted with COVID-19 or non-COVID-19 viral pneumonia), the way to evaluate its performance is to train it with some samples and then to check its predictions on others. The two groups of samples are called the **training set** and the **test set**, respectively. To ensure that the classifier does not "see" test samples in the training process, the two sets should be disjoint, i.e., their intersection is empty.

If we use a fixed pair of training and test sets to evaluate the classifier, it would be unclear how the same approach to train a classifier would remain effective on a different pair of training and test sets. To assess the robustness of a classifier configuration for a task, including the choice of the classifier family and the hyperparameters, cross validation (also called rotation estimation) [20] partitions all data into many non-overlapping sets, from which it forms different pairs (called **folds**) of training and test sets. In each fold, a one set is used for testing after all others are used for training. The test set rotates among all sets sequentially until all sets have been the test set exactly once. Finally, cross validation averages the performances across all folds.

All folds are independent but share the same classifier configuration. In each fold, the classifier is trained from scratch -- hence the classifier never sees test samples during the training stage. Because the training sets for all folds differ, they result in different classifiers. Note that the purpose of cross validation is to assess the robustness of using a classifier configuration to solve the problem, not to obtain the final classifier. Once a classifier configuration is found effective and robust, the final classifier is trained using all samples. The last step is usually called **refit** in many ML toolboxes.

To maximize the size of the training set, we use leave-one-out cross-validation (LOOCV) where each test set contains only one sample. One may imagine each fold of the LOOCV as a doctor using data from N-1 subjects to train the classifier and then applying the model on a new patient, while the entire process of LOOCV as N doctors doing this independently and we counting how many of them give thumbs up and thumbs down to the respective models. We can safely say that we test the approach on N subjects, and no prior knowledge of the test subject is "leaked" to the classifier.

**Grid search for hyperparameters**: As mentioned earlier, many classifier families contain hyperparameters, which are constants preset before training. Hyperparamters usually affect the performance of a classifier significantly and need to be customized for a problem. Grid search [21] is a technique to find the best hyperparameter, or hyperparameter combinations if multiple hyperparameters. For each hyperparameter, grid search first generates a set of candidates. Then it produces the Cartesian product for candidate sets of all hyperparameters, and exhaustively runs cross validation over all points in the Cartesian product [22]. The best cross validation performance is to be reported, and the point yielding it is used to produce the final model.

For SVMs, the hyperparameters to be searched include C for all soft-margin SVMs, and $\gamma$ additionally if using the RBF kernel. Two rounds of grid searches, first coarse and then fine, are performed. The grid search begins with a 2-based exponential range from $2^{-10}$ to $2^{10}$ and an initial step of 2 for C (and $\gamma$ when using the RBF kernel). In the second and the fine round, we linearly sample 20 points for C (and $\gamma$, respectively) in the range where the best C (and $\gamma$) is (are) found.

For DTs and RFs, the maximum depth (max_d in **Table 3**) is searched from 1 to 10 levels with a step of 1 level. For RFs, the number of trees (n_tree in **Table 3**) are searched for 10, 20, 50, and 100, and the number of features to consider when splitting nodes takes 3 values: total number of features, square root of it, and base-2 logarithm of it. All other hyperparameters for DTs and RFs use their respective default values in the scikit-learn toolbox [23].

**Evaluation metrics**: We use balanced accuracy, thus, average recall of both classes, as the primary metric. Other metrics to be reported include sensitivity (recall on positive class), specificity (recall on negative class), and precision [24]. In statistics, sensitivity (or specificity) means the ratio of true positive (or negative) to actual positive (or negative), i.e., sensitivity = $\frac{TP}{TP+FN}$ (specificity = $\frac{TN}{TN+FP}$ ). Precision means the ratio of true positive to predicted positive, i.e., precision = $\frac{TP}{TP+FP}$ .

In the primary and secondary tasks, COVID-19 samples are defined as positive. In the bonus task, severe COVID-19 is defined as positive. In the primary task, false positives are better than false negatives, because it is safer to isolate non-COVID-19 subjects seriously as COVID-19 subjects to cut the transmission chain than the other way around which spreads COVID-19 in the community. Hence, a classifier of a high sensitivity (and optionally a high precision) and low specificity is preferred over a classifier of low sensitivity and high specificity.

**Class weights**: Because the dataset is unbalanced, i.e., unequal amounts of samples in two classes, to get a fair evaluation, class weights are set inversely proportional to the number of samples in each class.

**Implementation details**: All experiments were carried out in scikit-learn [23]. In particular, the linear SVM and the RBF-kernel SVM are accessed through its wrapper for LIBLINEAR [25] and LIBSVM [26], respectively.

## 4.2 Quantitative results

Our initial effort shows very promising results. The performances of various classifiers on different classification tasks are given in **Table 3**.

**For the primary task**, moderate vs. viral (moderate COVID-19 vs. moderate non-COVID-19 viral pneumonia), we separately reporte the demographics-aware (first 4 rows) and demographics-free (rows 5 to 8) results. Overall, RBF-SVM and RF models alternate to take the first spot in both cases. When demographics are not considered, the RF-based classifier slightly outperforms RBF-SVM classifier with an accuracy of 85.68% vs 85.15%. When demographics are considered, the RF-based classifier loses edge slightly with an accuracy of 82.92% vs. 83.87%.

Thresholding-based classifiers (DTs and RFs) have quite different performances on the positive (moderate COVID-19) and negative (moderate non-COVID-19) classes. Take RFs as examples. The differences between sensitivity and specificity are 21.66% and 13.24%, respectively, with and without demographics. In contrast, SVM-based classifiers perform more balancedly between classes. For RBF-kernel SVMs, the sensitivity-specificity gaps are 7.27% and 8.77%, respectively, with and without demographics. The different sensitivity-specificity gaps of RFs and SVMs might be because that negative-class samples have a long-tail "diagonal" pattern that thresholding-based classifiers cannot effectively cope with as discussed in **Figure 3**. This is in line with our discussion in **Section 3.2**. In this sense, the RBF-kernel SVMs are superior to the RFs despite similar accuracies.

Between the two RBF-kernel SVMs, the degraphics-aware one is superior because it has both a high sensitivity (87.50%) and a high precision (91.46%), i.e., 87.50% moderate COVID-19 subjects are "caught" by the classifier and 91.46% of those caught by the classifier are actually contracted with COVID-19. The one without considering age and gender has a slightly higher precision (94.92%) but a much lower sensitivity (80.76% instead of 87.5%), meaning that nearly 1 out of 5 COVID-19 subjects are misclassified as non-COVID-19.

For all types of classifiers, the demographics-aware results are no worse than the demographics-free results in terms of sensitivity: 75% in both cases for linear SVMs, 87.5% vs. 81% for RBF-kernel SVMs, 93.75% vs. 92.31% for RFs, and 89.42% vs. 87.98% for DTs. This aligns well with reports that different age groups respond to COVID-19 differently and usually older groups are more vulnerable to COVID-19. The feature ranking analysis later in **Section 4.3** shows that gender plays a smaller role than age.

Hence, the conclusion for the primary task is that the RBF-kernel SVM considering age and gender is the best. Our web portal and released model is based on it. For the next two tasks, we focus on reporting the results of the two best classifier families, the RFs and the RBF-SVMs.

**Table 3**: Performances of different classifiers on different tasks.

| task | Classifier type | Best hyperparameters[2] | balanced accuracy (%) | sensitivity (%) | specificity (%) | precision (%) |
|---|---|---|---|---|---|---|
| **Primary**: moderate vs. viral (with age and gender) | SVM, linear kernel | C=1 | 78.78 | 75.00 | 82.56 | 91.23 |
| | SVM, RBF kernel | C=45, $\gamma$=0.0047 | **83.87** | 87.50 | 80.23 | 91.46 |
| | Random Forest | max_d = 9, n_tree = 50[3] | 82.92[2] | 93.75 | 72.09 | 89.41 |
| | Decision tree | max_d = 6 | 81.92 | 89.42 | 74.42 | 89.42 |
| **Primary**: moderate vs. viral (without age and gender) | SVM, linear kernel | C=3.01 | 78.2 | 75.00 | 81.40 | 90.70 |
| | SVM, RBF kernel | C=1.52, $\gamma$=0.00918 | 85.15 | 80.76 | 89.53 | 94.92 |
| | Random Forest | max_d=5, n_tree=20 | **85.68** | 92.31 | 79.07 | 91.43 |
| | Decision tree | max_d = 5 | 81.78 | 87.98 | 75.58 | 89.71 |
| **Secondary**: severe vs. viral (with age and gender) | SVM, RBF kernel | C=6.675, $\gamma$=0.0011 | **92.16** | 88.98 | 95.35 | 96.33 |
| | Random Forest | max_d=5, n_tree=50[2] | 86.35 | 88.98 | 93.72 | 88.24 |
| **Secondary**: severe vs. viral (without age and gender) | SVM, RBF kernel | C=24, $\gamma$=0.003397 | **89.04** | 83.89 | 94.18 | 95.19 |
| | Random Forest | max_d=3, n_tree=50[2] | 85.08 | 86.44 | 83.72 | 87.93 |
| **Bonus**: severe vs. moderate (with age and gender) | SVM, RBF kernel | C=3.67, $\gamma$=0.00335 | 76.39 | 75.21 | 77.56 | 65.67 |
| | Random Forest | max_d=5, n_tree = 20[2] | **76.44** | 67.52 | 85.37 | 72.48 |

**In the secondary task**: severe vs. viral (i.e., severe COVID-19 vs. moderate non-COVID-19 pneumonia), the RBF-SVMs outperform the RFs a lot, with balanced accuracies 92.16% vs. 86.35% when considering age and gender, and 89.04% vs. 85.08% when not. Unlike in the primary task where the accuracy differences between SVMs and RFs are within 0.5 percent point, in this task SVMs lead the RFs by 4 to 6 percent points. We hypothesize that this is due to the wide range of test values for severe subjects, causing extreme samples at "diagonal" long tails. This aligns with our explanation in **Section 3.2** that capturing the multidimensional distribution trend of data for a sloped decision hyperplane is better than thresholding on features. Like the case in the primary task, both types of classifiers do better when considering age and gender. Not only in terms of sensitivity but also in terms of accuracy.

---
[2] For all RFs, the optimal number of features is always the base-2 logarithm of the total number of features.
[3] depending on the random state

**For the onus task**: severe vs. moderate (i.e., severe COVID-19 vs. moderate COVID-19), we report the demographics-aware results only for space sake. It turns out to be the most difficult task, with both classifiers showing very low accuracies, 76.39% for the RBF-SVM and 76.44% for the RF. Hence classifying severe and moderate COVID-19 subjects from blood tests alone is not effective. Additional modalities, such as radiological findings or epidemiology information, need to be taken into consideration. However, remember that the bonus task was introduced out of curiosity rather than clinical needs.

## 4.3 Medical interpretation of the result

Science is not only about making things work but also why they work. Beyond the promising performances of machine learning models, we further want to know whether they capture the characteristics of COVID-19 that can be explained medically. One way to do so is to compute the feature importance, which measures how and to what extent a feature influences the classifiers. In the example in **Figure 2**, Feature 1 has a heavier influence on the decision hyperplane's slope than Feature 2, and we may say that Feature 1 is more important than Feature 2 for that toy task. If the importances of features can agree with medical explanations, then we are more sure on the effectiveness of using ML to solve our problem. For simplicity, we focus on two simple model families: the linear SVMs and the DTs here because there is a one-to-one correspondence between raw features and the elements of the classifiers (weights for linear SVMs or nodes for DTs). It is much harder to explain an RBF-kernel SVM or an RF. In this part of the study, we focus on the primary task: classifying moderate COVID-19 and moderate non-COVID-19 viral pneumonia from routine blood tests.

For a linear SVM, the more important a feature is, the (sloped) decision hyperplane will be more perpendicular to the axis of the feature, like Feature 1 in **Figure 2**, and its corresponding weight in the weight vector $W$ will have a higher magnitude. When the normalized weight on a feature is 1, the hyperplane is perpendicular to the axis of the feature, and the linear SVM becomes a comparator or a 1-level DT, like the vertical decision line in **Figure 2**. A normalized weight of a feature being 0 means that the feature has no influence on the classifier, e.g., Feature 2 has no influence on the vertical decision line.

For the DTs, we used Gini impurity to decide the splits of nodes. Hence, the importance score of a feature to a DT is its Gini impurity, which characterizes how well two classes can be separated by thresholding on this feature. Using the example in **Figure 1**, Feature 1 has a higher Gini impurity than Feature 2, because nearly all samples to the left of the decision line belong to one class and all to the right belong to the other. But a (non-sloped) decision line on Feature 2 will always result in a fair mixture of both classes. Hence, Feature 1 is more important than Feature 2.

The importance scores and ranks of features are listed in **Table 4**. Importance scores are the higher the better while ranks are the lower the better. For the SVM and the DT, CRP ranks at the top. CRP is a nonspecific marker of systemic inflammation, so it is not surprising to see it elevated in subjects with COVID (which is a systemic infection) than subjects without. The 0.509 normalized weight on CRP from the linear SVM shows that the slope of the hyperplane to the CRP axis is nearly 45 degrees, capturing a "diagonal" pattern illustrated in **Figure 2**.

The next few important features capture a clear impact of COVID-19 to our immune system. For the SVM, the next 3 features (ranks 2 to 4) are WBC count, neutrophil count, and lymphocyte count. For the DT, the 4th and 5th most important features are neutrophil percentage, and neutrophil count. WBCs, neutrophils, and lymphocytes are all cell categories in our immune system. Lymphocytes and neutrophils are two subclasses of WBCs fighting against viral and bacterial/fungal infections, respectively. Because all 4 features here are lower in the COVID-19 class, we hypothesize that our immune system is compromised more by COVID-19 than by other viral pneumonias. This is in line with reports that low WBC is found in 70% of COVID-19 subjects and lymphocytopenia in 63% [2].

Not surprisingly, the first organs attacked by COVID-19 are our lungs. The SVM and DT rank D-dimer at the 6th and 3rd spots, respectively, while the DT ranks platelet count at the 7th. D-dimer is a small protein fragment in the blood after a blood clot is degraded by fibrinolysis. The increase of D-dimer in blood for COVID-19 is probably due to more blood clots in lungs than non-COVID-19 viral infections. Platelets are also related to blood clots. The lower amount of platelets for COVID-19 is probably a result of consuming too many of them on forming clots. HGB is the protein responsible for carrying oxygen in red blood cells. It is the 6th most important feature per the DT. Its increase for COVID-19 could be a possible reaction to boost up the oxygen carrying function in our blood as clots in the lungs narrows the interface to air.

**Table 4**: Feature ranking according to two ML models, a linear SVM and a DT, and how their means change in COVID-19 with respect to non-COVID-19 pneumonia. Importance scores are the higher the better while ranks are the lower the better.

| feature ID | feature name | importance (rank) | | change in COVID-19 w.r.t. non-COVID-19 |
|---|---|---|---|---|
| | | by SVM | by DT | |
| 1 | WBC | 0.082 (2) | 0.008 (8) | lower |
| 2 | HGB | 0.015 (12) | 0.024 (6) | higher |
| 3 | platelet | 0.000 (15) | 0.023 (7) | lower |
| 4 | neutrophil % | 0.019 (11) | 0.050 (4) | lower |
| 5 | neutrophil # | 0.052 (4) | 0.027 (5) | lower |
| 6 | lymphocyte % | 0.047 (7) | 0.000 (9) | N/A |
| 7 | lymphocyte # | 0.076 (3) | 0.000 (10) | lower |
| 8 | CRP | 0.509 (1) | 0.638 (1) | higher |
| 9 | TBil | 0.051 (5) | 0.000 (11) | higher |
| 10 | BUN | 0.030 (9) | 0.000 (12) | lower |
| 11 | creatinine | 0.039 (8) | 0.143 (2) | higher |
| 12 | LDH | 0.027 (10) | 0.000 (13) | higher |
| 13 | D-dimer | 0.049 (6) | 0.087 (3) | higher |
| 14 | age | 0.003 (13) | 0.000 (14) | N/A |
| 15 | gender | 0.002 (14) | 0.000 (15) | N/A |

As we go down the ranks, the damage of COVID-19 spreads to the kidneys and the liver. SVM ranks TBL as the 6th most important feature. Increased TBL indicates damage to the kidney. In our data, COVID-19 subjects have higher TBL than non-COVID-19 subjects, showing possible more damage to the kidney caused by COVID-19 than by non-COVID-19 viral pneumonia. Although there is little report in literature about kidney damage to mild or moderate COVID-19 patients, it was found prevalent in patients who died from COVID-19 [27], with possible renal cell infections by the virus [28]. The 9th most important feature by SVM is BUN, which indicates how well both the kidneys and liver function. We observed decreased BUN in moderate COVID-19 patients than moderate non-COVID-19 patients. Because liver damage lowers the BUN level while kidney damage raises the BUN level, we hypothesize that COVID-19 causes severe damage to the liver, given that we already know that the kidneys are damaged badly from the rank (5th) of TBL per the SVM. Liver damage has been found prevalent in COVID-19 patients [29] without pre-existing liver conditions. Go back to the 8th most important feature by SVM, creatinine, which is also the 2nd most important feature by the DT. Creatinine is synthesized by the liver and removed from the blood by the kidneys. Because the analysis above hypothesizes liver damage which reduces its synthesis, we infer that creatinine increase for COVID-19 is due to the kidney damage.

The 10th most important feature per the SVM is LDH. Like CRP, it is also a nonspecific laboratory finding that is found in many different diseases. It is highly expressed in tissues throughout the body, and its elevation is often a marker of tissue damage. So in patients who are more septic (more severe infections) it would be higher. It is also in other contexts, such as liver failure, hemolysis (destruction of red blood cell), etc. Liver damage was discussed above. We hypothesize that the hemolysis is related to the high HGB levels in our data and the low blood oxygen symptom commonly seen on COVID-19 patients. Other study shows a very high level of LDH in >70% of COVID-19 subjects [2].

Lastly, both age and gender are ranked at or near the bottom for both classifiers, showing that age and gender play a little role in blood test changes due to COVID-19.

### 4.4 The web portal, source code and models

We believe openness as a core value of science. Hence we will release our results in 3 ways. First, we release a web portal for medical workers around the world to directly use our results in their clinical practices. The web portal is basically a web form with 15 boxes, corresponding to the 13 features plus age and gender. It can be opened on any computer or smartphone through a web browser. Entering the values for them and then clicking the submit button will generate a prediction, COVID-19 or not. To protect the privacy of the patients, no user data will leave the browser including being uploaded to our server. Our web portal serves only as a distributor to deliver the trained SVM model and parameters to the user's browser, and then all computation is done locally in the browser. The computer can even be disconnected from the Internet after the portal finishes loading. Or, the web portal can also be saved as a web page, copied to any computer or smartphones, and opened in any web browser. The web portal is developed in libsvm-js[4], which is compiled from LIBSVM using Emscripten, a transpiler from C++ to JavaScript. The web portal is at http://forrestbao.github.io/covid-19

The trained models, in both libsvm and python/sklearn format, are also released in the Github repository https://github.com/forrestbao/covid-19  Our source code is to be released upon the acceptance of this paper.

## 7. Conclusion and future work

In this paper, we proposed an approach to automatically separate COVID-19 and non-COVID-19 subjects both showing moderate pneumonia symptoms based on routine blood tests using machine learning. This approach is developed in response to the cost, availability, and throughout limits of other COVID-19 screening methods. We tested this approach on a dataset of 86 moderate non-COVID-19 viral pneumonia patients and 208 moderate COVID-19 patients. The initial result is promising with an RBF-kernel SVM achieving an 84% accuracy (class balanced), an 88% sensitivity, and a 92% precision. We further studied whether this approach can be used to separate severe COVID-19 against non-COVID-19 viral and to determine the stage of COVID-19. Our results can be explained from both machine learning and medical backgrounds. If validated on a larger population, our approach can provide another rapid COVID-19 testing solution that can be done using existing lab facilities capable of routine blood tests.

---

[4] https://github.com/mljs/libsvm

# Acknowledgments

Bao and He's work is partially supported by (U.S.) National Science Foundation (NSF) under grant numbers MCB-1821828 and CNS-1817089. Xu's work is partially supported by the National Science Foundation for Young Scientists of China (81703201), the Natural Science Foundation for Young Scientists of Jiangsu Province (BK20171076), the Jiangsu Provincial Medical Innovation Team (CXTDA2017029), the Jiangsu Provincial Medical Youth Talent program (QNRC2016548), the Jiangsu Preventive Medicine Association program (Y2018086), the Lifting Program of Jiangsu Provincial Scientific and Technological Association, and the Jiangsu Government Scholarship for Overseas Studies. Xinran Wang and Zeyu Yang helped build the web portal. Drs. Cen Chen and Minghui Qiu of Alibaba Group, China, also provided feedback on the experimental design. The authors would like to thank Tianyu Yan of Pennsylvania State University for her help on preparing the manuscript.

# References

[1] The Lancet Infectious Diseases, "COVID-19: endgames," *Lancet Infect. Dis.*, Apr. 2020, doi: 10.1016/S1473-3099(20)30298-X.
[2] C. Huang *et al.*, "Clinical features of patients infected with 2019 novel coronavirus in Wuhan, China," *Lancet*, vol. 395, no. 10223, pp. 497–506, Feb. 2020.
[3] F. He, Y. Deng, and W. Li, "Coronavirus disease 2019: What we know?," *J. Med. Virol.*, Mar. 2020, doi: 10.1002/jmv.25766.
[4] H. X. Bai *et al.*, "Performance of radiologists in differentiating COVID-19 from viral pneumonia on chest CT," *Radiology*, p. 200823, Mar. 2020.
[5] L. L. Maragakis, "Coronavirus Disease 2019 vs. the Flu," *Johns Hopkins Medicine*. https://www.hopkinsmedicine.org/health/conditions-and-diseases/coronavirus/coronavirus-disease-2019-vs-the-flu (accessed Apr. 30, 2020).
[6] J. M. Sharfstein, S. J. Becker, and M. M. Mello, "Diagnostic Testing for the Novel Coronavirus," *JAMA*, Mar. 2020, doi: 10.1001/jama.2020.3864.
[7] A. Harmon, "Why We Don't Know the True Death Rate for Covid-19," *New York TImes*, Apr. 17, 2020.
[8] National Heart, Lung, and Blood Institute, "Blood tests." https://www.nhlbi.nih.gov/health-topics/blood-tests (accessed Apr. 30, 2020).
[9] L. Breiman, "Random Forests," *Mach. Learn.*, vol. 45, no. 1, pp. 5–32, Oct. 2001.
[10] C. Cortes and V. Vapnik, "Support-vector networks," *Mach. Learn.*, vol. 20, no. 3, pp. 273–297, Sep. 1995.
[11] M. Fernández-Delgado, E. Cernadas, S. Barro, and D. Amorim, "Do we Need Hundreds of Classifiers to Solve Real World Classification Problems?," *J. Mach. Learn. Res.*, vol. 15, no. 90, pp. 3133–3181, 2014, Accessed: May 03, 2020. [Online].
[12] M. Wainberg, B. Alipanahi, and B. J. Frey, "Are Random Forests Truly the Best Classifiers?," *J. Mach. Learn. Res.*, vol. 17, no. 110, pp. 1–5, 2016, Accessed: May 03, 2020. [Online].
[13] S. Sathe and C. C. Aggarwal, "Nearest Neighbor Classifiers Versus Random Forests and Support Vector Machines," in *2019 IEEE International Conference on Data Mining (ICDM)*, Nov. 2019, pp. 1300–1305.
[14] National Health Commission of China, English translation by WHO China office, "Diagnosis and Treatment Protocol for Novel Coronavirus Pneumonia (Trial version 7)," Mar. 2020. [Online]. Available: https://www.chinadaily.com.cn/pdf/2020/1.Clinical.Protocols.for.the.Diagnosis.and.Treatment.of.COVID-19.V7.pdf.
[15] N. Chen *et al.*, "Epidemiological and clinical characteristics of 99 cases of 2019 novel coronavirus

pneumonia in Wuhan, China: a descriptive study," *Lancet*, vol. 395, no. 10223, pp. 507–513, Feb. 2020.

[16] L. Breiman, J. Friedman, C. J. Stone, and R. A. Olshen, *Classification and regression trees*. CRC press, 1984.

[17] L. E. Raileanu and K. Stoffel, "Theoretical Comparison between the Gini Index and Information Gain Criteria," *Ann. Math. Artif. Intell.*, vol. 41, no. 1, pp. 77–93, May 2004.

[18] U. M. Fayyad and K. B. Irani, "The Attribute Selection Problem in Decision Tree Generation," in *Proceedings of the Tenth National Conference on Artificial Intelligence (AAAI'92)*, 1992, pp. 104–110.

[19] B. E. Boser, I. M. Guyon, and V. N. Vapnik, "A training algorithm for optimal margin classifiers," in *COLT '92: Proceedings of the fifth annual workshop on Computational Learning Theory*, 1992, pp. 144–152.

[20] R. Kohavi, "A study of cross-validation and bootstrap for accuracy estimation and model selection," in *IJCAI'95: Proceedings of the 14th international joint conference on Artificial intelligence*, 1995, vol. 14, pp. 1137–1145.

[21] C. wei Hsu, C. chung Chang, and C. jen Lin, "A practical guide to support vector classification," *National Taiwan University, Taiwan, Tech. Rep*, 2010.

[22] J. Bergstra and Y. Bengio, "Random Search for Hyper-Parameter Optimization," *J. Mach. Learn. Res.*, vol. 13, no. 10, pp. 281–305, 2012, Accessed: May 09, 2020. [Online].

[23] F. Pedregosa *et al.*, "Scikit-learn: Machine Learning in Python ," *J. Mach. Learn. Res.*, vol. 12, pp. 2825–2830, 2011.

[24] J. Davis and M. Goadrich, "The relationship between Precision-Recall and ROC curves," in *Proceedings of the 23rd international conference on Machine learning*, Pittsburgh, Pennsylvania, USA, Jun. 2006, pp. 233–240, Accessed: Apr. 30, 2020. [Online].

[25] R.-E. Fan, K.-W. Chang, C.-J. Hsieh, X.-R. Wang, and C.-J. Lin, "LIBLINEAR: A Library for Large Linear Classification," *J. Mach. Learn. Res.*, vol. 9, no. Aug, pp. 1871–1874, 2008, Accessed: Apr. 30, 2020. [Online].

[26] C.-C. Chang and C.-J. Lin, "LIBSVM: A library for support vector machines," *ACM Transactions on Intelligent Systems and Technology*, vol. 2, pp. 27:1–27:27, 2011.

[27] Y. Cheng *et al.*, "Kidney disease is associated with in-hospital death of patients with COVID-19," *Kidney Int.*, vol. 97, no. 5, pp. 829–838, May 2020.

[28] H. Su *et al.*, "Renal histopathological analysis of 26 postmortem findings of patients with COVID-19 in China," *Kidney Int.*, Apr. 2020, doi: 10.1016/j.kint.2020.04.003.

[29] C. Zhang, L. Shi, and F.-S. Wang, "Liver injury in COVID-19: management and challenges," *Lancet Gastroenterol Hepatol*, vol. 5, no. 5, pp. 428–430, May 2020.